\documentclass[letterpaper]{article} 
\usepackage{aaai23}  
\usepackage{times}  
\usepackage{helvet}  
\usepackage{courier}  
\usepackage[hyphens]{url}  
\usepackage{graphicx} 
\usepackage{natbib}  
\usepackage{caption} 
\usepackage{algorithm}
\usepackage{algorithmic}

\usepackage{newfloat}
\usepackage{listings}
\DeclareCaptionStyle{ruled}{labelfont=normalfont,labelsep=colon,strut=off} 
\floatstyle{ruled}
\newfloat{listing}{tb}{lst}{}
\floatname{listing}{Listing}
\usepackage{amsmath}
\usepackage{amssymb}
\usepackage{booktabs}
\usepackage{multirow}
\usepackage{array}
\usepackage{subcaption}
\usepackage{makecell}
\usepackage{hyperref}

\title{TopicFM: Robust and Interpretable Topic-Assisted Feature Matching}
\author {
    Khang Truong Giang \textsuperscript{\rm a},
    Soohwan Song \textsuperscript{\rm b}\footnote{Corresponding authors},
    Sungho Jo \textsuperscript{\rm a}\footnotemark[\value{footnote}]
}
\affiliations {
    \textsuperscript{\rm a} School of Computing, KAIST, Daejeon 34141, Republic of Korea\\
    \textsuperscript{\rm b} Intelligent Robotics Research Division, ETRI, Daejeon 34129, Republic of Korea\\
    khangtg@kaist.ac.kr, soohwansong@etri.re.kr, shjo@kaist.ac.kr
}

\usepackage{bibentry}

\begin{document}

\maketitle

\begin{abstract}
This study addresses an image-matching problem in challenging cases, such as large scene variations or textureless scenes. To gain robustness to such situations, most previous studies have attempted to encode the global contexts of a scene via graph neural networks or transformers. However, these contexts do not explicitly represent high-level contextual information, such as structural shapes or semantic instances; therefore, the encoded features are still not sufficiently discriminative in challenging scenes. We propose a novel image-matching method that applies a topic-modeling strategy to encode high-level contexts in images. The proposed method trains latent semantic instances called topics. It explicitly models an image as a multinomial distribution of topics, and then performs probabilistic feature matching. This approach improves the robustness of matching by focusing on the same semantic areas between the images. In addition, the inferred topics provide interpretability for matching the results, making our method explainable. Extensive experiments on outdoor and indoor datasets show that our method outperforms other state-of-the-art methods, particularly in challenging cases. Our code is available at \href{https://github.com/TruongKhang/TopicFM}{github}.
\end{abstract}

\section{Introduction}
\label{sec:intro}

\begin{figure}[t]
\centering
\includegraphics[scale=0.26]{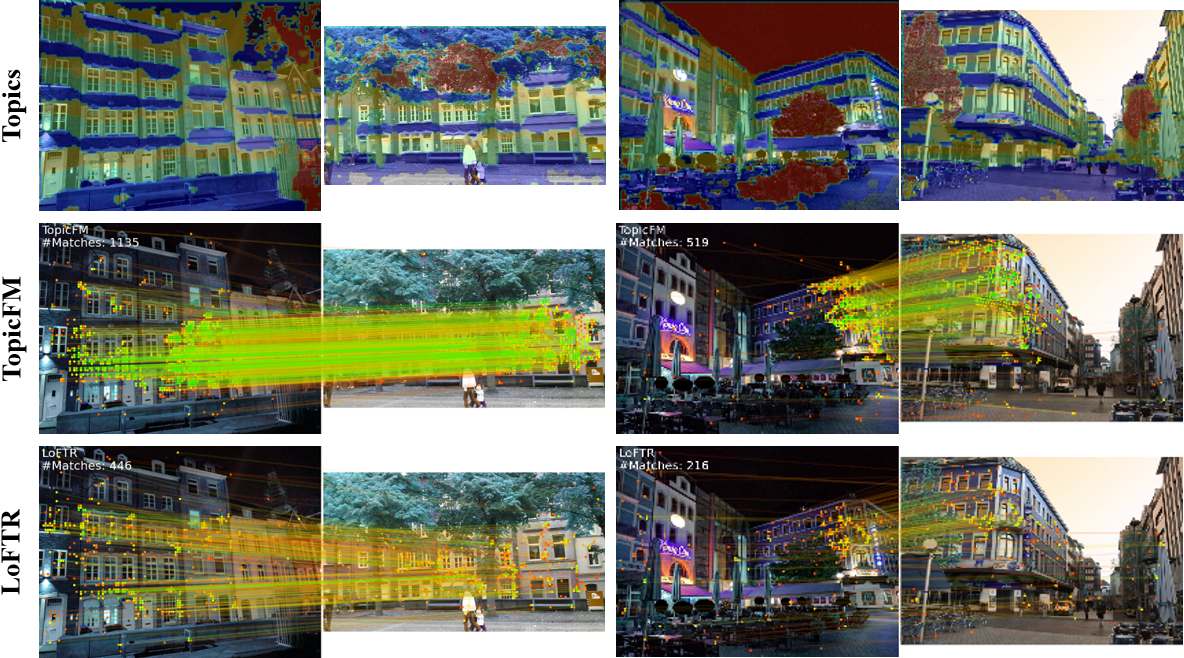}
\caption{The main idea of our human-friendly topic-assisted feature matching, \textbf{TopicFM}. This method represents an image as a set of topics marked in different colors and quickly recognizes the same structures between an image pair. It then leverages the distinctive information of each topic to augment the pixel-level representation. As shown in the comparing illustrations above, TopicFM provides robust and accurate matching results, even for challenging scenes with large illumination and viewpoint variations.}
\label{fig:intro}
\end{figure}

Image matching is a long-standing problem in computer vision. It aims to find pixel-to-pixel correspondences across two or more images. Conventional image matching methods \cite{lowe2004distinctive,bay2008speeded,sattler2012improving} 
usually involve the following steps: i) local feature detection, ii) feature description, iii) matching, and iv) outlier rejection. These methods usually involve extracting sparse handcrafted local features (i.e., SIFT \cite{lowe2004distinctive}, SURF \cite{bay2008speeded}, or ORB \cite{rublee2011orb}) and matching them using a nearest neighbor search. Many recent studies 
have adopted convolutional neural networks (CNNs) to extract local features, which significantly outperform the conventional handcrafted features. However, such methods sometimes fail in challenging cases, such as illumination variations, repetitive structures, or low-texture conditions.

To address this issue, detector-free methods \cite{li2020dual,rocco2018neighbourhood} have been proposed. These methods estimate dense feature maps without feature detection and perform pixel-wise dense matching. Furthermore, a coarse-to-fine strategy has been applied to improve the computational efficiency. The strategy finds matches at a coarse level, and then refines the matches at a finer level. Such methods 
\cite{sun2021loftr,wang2022matchformer} produce a large number of matches, even for repetitive patterns and texture-less scenes, thus achieving state-of-the-art performance.

However, detector-free methods still have some factors that degrade the matching performance. First, these methods cannot adequately incorporate the global context of a scene for feature matching. Several methods have attempted to implicitly capture global contextual information via transformers \cite{sun2021loftr,jiang2021cotr} 
or patch-level matches \cite{zhou2021patch2pix}, but higher-level contexts, such as semantic instances, should be effectively exploited to learn robust representations. Second, they exhaustively search for all features of the entire image area. Therefore, their matching performance is considerably low when there are limited covisible regions between images. Finally, these methods require intensive computation of dense matching, which increases runtime. Therefore, a more efficient model is needed for real-time applications such as SLAM \cite{mur2015orb}.

In this study, we propose a novel detector-free feature matching method, TopicFM, that encodes high-level contextual information on images based on a topic modeling strategy in data mining \cite{blei2003latent,yan2013biterm}. TopicFM models an image as a multinomial distribution over topics, where a topic represents a latent semantic instance such as an object or structural shape. TopicFM then performs probabilistic feature matching based on the distribution of the latent topics. It integrates topic information into local visual features to enhance their distinctiveness. Furthermore, it effectively matches features within overlapping regions between an image pair by estimating the covisible topics. Therefore, TopicFM provides robust and accurate feature-matching results, even for challenging scenes with large scale and viewpoint variations.

The proposed method also provides interpretability for matching results across topics. Fig. \ref{fig:intro} illustrates the representative topics inferred from the image matching results. In Fig. \ref{fig:intro}, the image regions with the same object or structure are assigned to the same topic. Based on the sufficient high-level context information in the topic, TopicFM can learn discriminative features. Therefore, it is able to find the accurate dense correspondences in the same topic regions. This approach is similar to the human cognitive system, in which humans quickly recognize covisible regions based on semantic information and then search for matching points in these regions. By applying this top-down approach, our method successfully detected dense matching points in various challenging image conditions.

Going one step further, we designed an efficient end-to-end network architecture to accelerate the computation. We adopted a coarse-to-fine framework and constructed lightweight networks for each stage. In particular, TopicFM only focuses on the same semantic areas between the images for learning features. Therefore, our method requires less computation compared to other methods \cite{sarlin2020superglue,sun2021loftr,wang2022matchformer} that apply the transformer to the whole domain.

The contributions of this study are as follows:
\begin{itemize}
    \item We present a novel feature-matching method that fuses local context and high-level semantic information into latent features using a topic modeling strategy. This method produces accurate dense matches in challenging scenes by inferring covisible topics.
    \item We formulate the topic inference process as a learnable transformer module. These inferred topics can provide interpretability for matching results with humans.
    \item We design an efficient end-to-end network model to achieve real-time performance. This model processes image frames much faster than state-of-the-art methods such as Patch2Pix \citep{zhou2021patch2pix} and LoFTR \citep{sun2021loftr}.
    \item We empirically evaluate the proposed method through extensive experiments. We also provide results on the interpretability of our topic models. Source code for the proposed method is publicly available.
\end{itemize}

\section{Related Works}
\label{sec:related_works}
\subsubsection{Image Matching}
The standard pipeline for image matching \cite{ma2021image} consists of four steps: feature detection, description, and matching, and outlier rejection. Traditional feature detection-and-description methods such as SIFT \cite{lowe2004distinctive}, SURF \cite{bay2008speeded}, 
and BRIEF \cite{calonder2010brief}, although widely used in many applications, require a complicated selection of hyperparameters to achieve reliable performance \cite{efe2021effect}. Twelve years after SIFT, a fully learning-based architecture, LIFT \cite{yi2016lift}, was proposed to address the hand-crafting issue of traditional approaches. Many studies \cite{detone2018superpoint,ono2018lf,dusmanu2019d2,revaud2019r2d2,bhowmik2020reinforced,tyszkiewicz2020disk} also proposed learning-based approaches, which have become dominant in feature detection and description. However, their methods mainly adopt standard CNNs to learn features from local context information, which is less effective when processing low-textured images.

To address this issue, some studies \cite{sun2021loftr,wang2022matchformer,luo2019contextdesc,luo2020aslfeat} have additionally considered global context information. ContextDesc \cite{luo2019contextdesc} and ALSFeat \cite{luo2020aslfeat} proposed a geometric context encoder using a large patch sampler and deformable CNN, respectively. LoFTR \cite{sun2021loftr} applies transformers with self- and cross-attentions to extract dense feature maps. Although these methods are technically sound, they are unable to encode high-level contexts such as objects or structural shapes. They cannot explicitly represent hidden semantic structures in an image and lack interpretability. However, our method can capture latent semantic information via a topic modeling strategy; therefore, our matching results would be fairly interpretable.

Given two sets of features produced by the detection-and-description methods, a basic feature-matching algorithm applies the nearest neighbor search \cite{muja2014scalable} or ratio test \cite{lowe2004distinctive} to find potential correspondences. Next, the matching outliers are rejected by RANSAC \cite{fischler1981random}
, consensus- or motion-based heuristics \cite{lin2017code,bian2017gms}, 
or learning-based methods \cite{yi2018learning,zhang2019learning}. 
The outlier rejection performance relies heavily on the accuracy of the trained features. Recently, several studies \cite{sarlin2020superglue,chen2021learning,shi2022clustergnn} employed an attentional graph neural network (GNN) to enhance the quality of extracted features. These features were matched with an optimal transport layer \cite{cuturi2013sinkhorn}. 
As the performance of these methods depends on the features of the detector, these methods cannot guarantee robust and reliable performance.

Motivated by the above observation, several studies \cite{zhou2021patch2pix,sun2021loftr,wang2022matchformer,jiang2021cotr} have proposed an end-to-end network architecture that performs image matching in a single forward pass instead of dividing separate steps. The network directly processed dense feature maps instead of extracting sparse feature points. Several studies applied a coarse-to-fine strategy to process the dense features of a high-resolution image efficiently. Patch2Pix 
detects coarse matches in low-resolution images and gradually refines them at higher resolutions. Similarly, other coarse-to-fine methods \cite{sun2021loftr,jiang2021cotr,wang2022matchformer} learn robust and distinctive features using transformers and achieve state-of-the-art performance. However, these methods remain inefficient when propagating global context information to the entire image region. We argue that the invisible regions between an image pair are redundant and may cause noise when learning the features with transformers. Therefore, we propose a topic modeling approach to utilize adequate context cues for learning representations.

\subsubsection{Interpretable Image Matching}
The interpretability of vision models has recently been actively researched \cite{zhou2016learning,selvaraju2017grad,bau2018gan,chefer2021transformer}. It aims to explain a certain decision or prediction in image recognition 
\cite{williford2020explainable,wang2021interpretable}, or deep metric learning \cite{zhao2021towards}. In image matching, the detector-based methods \cite{forstner2009detecting,lowe2004distinctive} can estimate interpretable feature keypoints such as corners, blobs, or ridges. However, detected features do not represent spatial or semantic structures. Otherwise, existing end-to-end methods only extract dense feature maps using the local context via CNNs \cite{zhou2021patch2pix} or global context via transformers \cite{sun2021loftr,wang2022matchformer}. However, these approaches cannot explicitly describe the details of the observed context information; therefore, their results lack interpretability.

The human cognitive system quickly recognizes covisible regions based on high-level contextual information, such as objects or structures. It then determines the matching points in the covisible regions. Inspired by this cognitive process, we designed an end-to-end model that is human-friendly. It categorizes local structures in images into different topics and uses only the information within topics to augment features. Moreover, our method performs interpretable matching by selecting important topics in the covisible regions of the two images. To the best of our knowledge, our method is the first to explicitly introduce interpretability to an image matching task.

\subsubsection{Semantic Segmentation} 
Various deep learning models for semantic segmentation have been introduced, such as fully convolutional networks \cite{long2015fully}, encoder-decoder \cite{yuan2020object}, R-CNN-based \cite{he2017mask}, or attention-based models \cite{strudel2021segmenter}. Unlike semantic segmentation, our topic modeling does not strictly detect semantic objects. However, it can effectively exploit the local structures or shapes, which benefits learning pixel-level representation for feature matching. Moreover, the topics can be trained in a self-supervised manner without requiring a large amount of labeled training data, as in semantic segmentation.

\begin{figure*}[ht]
\centering
\includegraphics[scale=0.33]{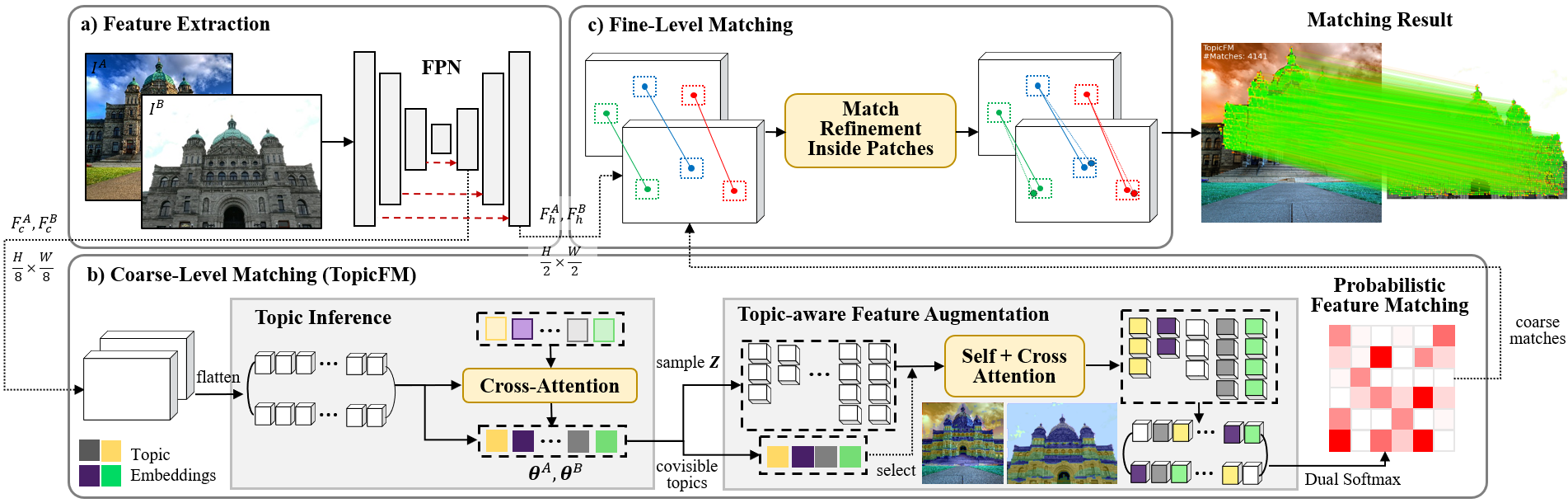}
\caption{
Overview of the proposed architecture. (a) Our method first extracts multilevel feature maps. (b) Next, the method finds coarse matches from low-resolution features. It infers a topic distribution via a cross-attention layer with topic embeddings. It then samples topic labels of each feature point and augments the features with self/cross attention layers. The coarse matches are determined by estimating a matching probability with dual-softmax. (c) Finally, our method refines the coordinates inside the cropped patches at high resolution.
}
\label{fig:architecture}
\end{figure*}

\section{Proposed Method}
\label{sec:method}
\subsection{Coarse-to-fine Architecture}
\label{sec:method_overview}

This study addresses the feature-matching problem of an image pair. Let $F^A$ and $F^B$ be the feature maps extracted from images $I^A$ and $I^B$, respectively. Our objective is to find accurate and dense matching correspondences between two feature points, $f_i^A \in F^A$ and $f_j^B \in F^B$. We employ a coarse-to-fine architecture \cite{sun2021loftr} that trains a feature-matching network end-to-end. This architecture estimates coarse matches from low-resolution features and refines the matches to a finer level. This approach makes it possible to perform feature matching of high-resolution images in real time while preserving the pixel-level accuracy. 

Fig. 2 depicts the proposed architecture for feature matching, which is composed of three steps: i) feature extraction, ii) coarse-level matching, and iii) fine-level refinement. The feature extraction step generates multiscale dense features through a UNet-like architecture \cite{lin2017feature}. Let $\{F_c^A,F_c^B\}$ and $\{F_f^A,F_f^B\}$ be pairs of coarse- and finer-lever feature maps of an image pair $\{I^A,I^B\}$, respectively. The coarse matching method estimates the matching probability distribution of $\{F_c^A,F_c^B\}$ using a topic-assisted matching module, TopicFM. It then determines coarse correspondences based on the probability distribution (see the next section). The last stage refines the coarse matches to a finer level with high-resolution features $\{F_f^A,F_f^B\}$. We adopted the matching refinement method of LoFTR directly \cite{sun2021loftr}. For each coarse match $(i,j)$, the method finds the best matching coordinate in $F_f^B$ by measuring the similarities between a feature point $F_{f,i}^A \in F_f^A$ for all features of the cropped patch at $F_{f,j}^B \in F_f^B$.

\subsection{Topic-assisted Feature Matching}
\label{sec:method_topicfm}
\subsubsection{Probabilistic Feature Matching}
The coarse feature maps ${F_c^A,F_c^B}$ can be regarded as a bag-of-visual-words \cite{sivic2003video,csurka2004visual}, where each feature vector represents a visual word. Let $m_{ij}$ be a random variable that indicates an event in which the $i^{th}$ feature $F_{c,i}^A$ is matched to the $j^{th}$ feature $F_{c,j}^B$. 
Given two feature sets $\{F_c^A,F_c^B\}$, our goal is to estimate the match distribution of all possible matches $M = \{ m_{ij} \}$ \cite{bhowmik2020reinforced}:

\begin{equation}
\label{eq:prob_feat_match}
    P(M \mid F_c^A,F_c^B ) = \prod_{m_{ij} \in M} P \left( m_{ij} |F_c^A,F_c^B \right)
\end{equation}
The matches with high match probability $P \left( m_{ij} |F_c^A,F_c^B \right)$ are selected as the coarse correspondences.
Existing methods \cite{bhowmik2020reinforced,sun2021loftr,sarlin2020superglue} directly infer the matching probabilities using Softmax \cite{bhowmik2020reinforced}, Dual-Softmax \cite{sun2021loftr}, or optimal transport with Signkhorn regularization \cite{sarlin2020superglue}. Unlike these methods, TopicFM incorporates the latent distribution of topics to estimate the matching distribution.

%
To solve the matching problem of Eq. \ref{eq:prob_feat_match}, our method infers a topic distribution for each feature point (Eq. \ref{eq:topic_infer_norm}). It then estimates a matching probability conditioned on topics for each matching candidate (Eq. \ref{eq:log_likelihood}). A sampling strategy is employed to calculate this probability (Eq. \ref{eq:conditional_match_dis} and Eq. \ref{eq:topic_sampling}).  Finally, our method selects the coarse matches from the candidates using probability thresholding.

\subsubsection{Topic Inference via Transformers}

We assume that the structural shapes or semantic instances of the images in a specific dataset can be categorized into $K$ topics. Therefore, each image can be modeled as a multinomial distribution over $K$ topics. The probability distribution of the topics was assigned to each feature point. 

Let $z_i$ and $\theta_i$ be a topic indicator and topic distribution for feature $F_i$, respectively, where $z_i \in \{1,…,K\}$ and $\theta_{i,k} = p \left( z_i = k \mid F \right)$ are the probabilities for assigning $F_i$ to topic $k$. We represent topic k as an embedding vector, $T_k$, which is trainable. To estimate $\theta_i$, our method infers the local topic representations $\Hat{T}_k$ from the global representations $T_k$ using transformers:
\begin{equation}
\label{eq:topic_infer_trans}
    \Hat{T}_k = \mathcal{CA}(T_k, F)
\end{equation}
where $\mathcal{CA}(T_k,F)$ is the cross-attention layer between queries $T_k$, keys $F$, and values $F$. This function collects relevant information from an image of each topic. Finally, the topic probability $\theta_{i, k}$ is defined as the distance between feature $F_i$ and individual topics $\Hat{T}_k$ as follows:
\begin{equation}
\label{eq:topic_infer_norm}
    \theta_{i, k} = \frac{\langle \hat{T}_k, F_i \rangle}{\sum_{h=1}^K \langle \hat{T}_h, F_i \rangle}
\end{equation}

\subsubsection{Topic-aware Feature Augmentation}
This section describes the computation of Eq. \ref{eq:prob_feat_match} using inferred topics. We augment the features based on the high-level contexts of topics to enhance their distinctiveness. The augmented features are then used to estimate matching probability more precisely. Given a feature point pair $(F_{c,i}^A,F_{c,j}^B)$, we define an assigned topic of $z_{ij}$ as a random variable $z_{ij} \in \mathcal{Z} = \{ 1,2,…,K,NaN \}$. If $z_{ij} = k$ ($k = 1,…,K$), the pair belongs to the same topic $k$. Otherwise, $z_{ij} = NaN$ indicates that $F_{c,i}^A$ and $F_{c,j}^B$ do not belong to the same topic; therefore, they are highly unmatchable.

We define $z_{ij}$ as a latent variable for computing the matching distribution in Eq. \ref{eq:prob_feat_match} as follows:
\begin{multline}
\label{eq:log_likelihood}
    \log P\left(M \mid F_c^A, F_c^B \right) = \sum_{m_{ij} \in M} \log P  \left(m_{ij} \mid F_c^A, F_c^B \right) \\
    = \sum_{m_{ij} \in M} \log \sum_{k \in \mathcal{Z}} P \left(m_{ij}, z_{ij} = k \mid F_c^A, F_c^B \right)
\end{multline}
To compute Eq. \ref{eq:log_likelihood}, we approximated this equation with an evidence lower bound (ELBO):

\begin{multline}
\label{eq:lower_bound}
    \mathcal{L}_{ELBO} 
    = \sum_{m_{ij}} \sum_{k \in \mathcal{Z}} P\left(z_{ij} = k \mid F_c \right) \log P\left(m_{ij} \mid z_{ij}, F_c \right) \\
    = \sum_{m_{ij}} E_{p(z_{ij})}  \log P\left(m_{ij} \mid z_{ij}, F_c^A, F_c^B\right)
\end{multline}
where $P\left(m_{ij} |z_{ij},F_c^A,F_c^B\right)$ refers to the matching probability conditioned on topic $z_{ij}$.  Eq. \ref{eq:lower_bound} can be estimated by applying Monte-Carlo (MC) sampling, as follows: 
\begin{gather}
\label{eq:conditional_match_dis}
    \mathcal{L}_{ELBO} = \sum_{m_{ij}  \in M} \frac{1}{S} \sum_{s=1}^S \log P\left(m_{ij} \mid z_{ij}^{(s)}, F_c^A, F_c^B \right) \\
    z_{ij}^{(s)} \sim P\left(z_{ij} \mid F_c^A, F_c^B \right)
\label{eq:topic_sampling}
\end{gather}
where $S$ is the number of samples $(S \ll K)$. This sampling approach improves computational efficiency because it is unnecessary to iterate all $K$ topics to compute the expectation in Eq. \ref{eq:lower_bound}. Finally, the problem is reduced to the computation of the topic distribution $P(z_{ij} \mid F_c^A, F_c^B)$ and the conditional matching distribution $P(m_{ij} \mid z_{ij}^{(s)}, F_c^A, F_c^B)$.

\paragraph{\textit{Topic Distribution}}
We estimate the distribution of $z_{ij}$ by factorizing it into two distributions of $z_i$ and $z_j$ as follows:
\begin{multline}
\label{eq:topic_dis_k}
    P(z_{ij} = k \mid F_c^A, F_c^B ) = \\ P(z_i = k \mid F_c^A )  P(z_j = k \mid F_c^B)
    = \theta_{i,k}^A \theta_{j,k}^B
\end{multline}
where $\theta_{i,k}^A$, $\theta_{j,k}^B$ are computed using Eq. \ref{eq:topic_infer_trans} and Eq. \ref{eq:topic_infer_norm}. This represents the probability of assigning feature pair $\{F_{c,i}^A, F_{c,j}^B\}$ to a specific topic $k \in \{1,…,K\}$. The probability of being in at least one topic is calculated as follows:
\begin{equation}
\label{eq:topic_dis_all_k}
    P(z_{ij} \in \{1,…,K\} \mid F_c^A,F_c^B) = \sum_{k=1}^K \theta_{i,k}^A \theta_{j,k}^B
\end{equation}
Otherwise, the probability of not being on the same topic is calculated by
\begin{equation}
\label{eq:topic_dis_nan}
\begin{split}
    P(z_{ij} = NaN \mid .) &= 1 - \sum_{k=1}^K P(z_{ij} = k \mid .) \\
    &= 1 - \sum_{k=1}^K \theta_{i,k}^A \theta_{j,k}^B
\end{split}
\end{equation}
In summary, the topic distribution for each pair of features was determined as follows:
\begin{equation}
    P(z_{ij} = k \mid F_c) = \begin{cases} \theta_{i,k}^A \theta_{j,k}^B & k \in \{1…K\} \\ 1 - \sum_{k=1}^K \theta_{i,k}^A \theta_{j,k}^B  & k = NaN \end{cases}
\end{equation}
We can sample $z_{ij}^{(s)}$ from this distribution by sampling $z_i^{(s)}$ and $z_j^{(s)}$ from $\theta_i^A$ and $\theta_j^B$ separately based on the independent and identically distributed (i.i.d.) assumption:
\begin{equation}
    z_{ij}^{(s)} = \begin{cases} k & \text{if} \quad z_i^{(s)} = z_j^{(s)} = k \\ NaN & \text{if} \quad z_i^{(s)} \ne z_j^{(s)} \end{cases}
\end{equation}

\paragraph{\textit{Conditional Matching Distribution}}
After sampling, we classified a pair of features into topics. Let $F_c^{A, \Tilde{k}} \subset F_c^A$ and $F_c^{B, \Tilde{k}} \subset F_c^B$ be a set of features sampled with topic $\Tilde{k} = z_{ij}^{(s)}$. These features are augmented to improve their distinctiveness by applying self- and cross-attentions (SA and CA) of the transformer  \cite{sarlin2020superglue,sun2021loftr}:
\begin{equation*}
    \hat{F}_{c,i}^{A, \Tilde{k}} \leftarrow \mathcal{SA} \left( F_{c,i}^{A, \Tilde{k}}, F_c^{A, \Tilde{k}} \right), \quad 
    \hat{F}_{c,j}^{B, \Tilde{k}} \leftarrow \mathcal{SA} \left( F_{c,j}^{B, \Tilde{k}}, F_c^{B, \Tilde{k}} \right)
\end{equation*}
\begin{equation*}
    \hat{F}_{c,i}^{A, \Tilde{k}} \leftarrow \mathcal{CA} \left( F_{c,i}^{A, \Tilde{k}}, F_c^{B, \Tilde{k}} \right), \quad 
    \hat{F}_{c,j}^{B, \Tilde{k}} \leftarrow \mathcal{CA} \left( F_{c,j}^{B, \Tilde{k}}, F_c^{A, \Tilde{k}} \right)
\end{equation*}
This augmentation learns powerful representation by considering adequate context information inside the topic $\Tilde{k}$. Finally, the matching probability conditioned on topic $z_{ij}^{(s)}$ in Eq. \ref{eq:conditional_match_dis} is determined by computing the feature distance and normalizing it with a dual-softmax \cite{sun2021loftr}:
\begin{equation}
    P(m_{ij} \mid z_{ij}^{(s)}=\Tilde{k}, F_c^A, F_c^B) = DS\left( \langle \hat{F}_{c,i}^{A,\Tilde{k}}, \hat{F}_{c,j}^{B, \Tilde{k}} \rangle \right)
\end{equation}

To reduce redundant computation, we only augmented the features with covisible topics. Covisible topics were determined by comparing the topic distributions of the two images. The topic distribution in an image is estimated by aggregating the distributions of all features: 
\begin{equation}
    \theta_k^A \propto \sum_{i=1}^{|F_c^A|} \theta_{i,k}^A, \quad \theta_k^B \propto \sum_{j=1}^{|F_c^B|} \theta_{j,k}^B
\end{equation}
where $\propto$ denotes the normalization operator. We then calculated the covisible probability by multiplying the two topic distributions as $\theta_{k}^{Vis} = \theta_{k}^A \theta_{k}^B$. Finally, the most important topics were selected as the covisible topics for feature augmentation based on probability.

\subsection{Implementation Details}

\subsubsection{Efficient Model Design} To achieve a fast computation, we designed an efficient lightweight network for each coarse-to-fine step. For feature extraction, we applied a standard UNet instead of ResUnet, as in other methods \cite{zhou2021patch2pix,sun2021loftr}. In the coarse matching step, TopicFM uses a single block of self/cross-attention and shares it across topics to extract the features. This operation is applied only to covisible topics; therefore, it is more efficient than methods that use a multi-block transformer \cite{sun2021loftr,wang2022matchformer}. Finally, in the fine matching step, our method applies only a cross-attention layer instead of both self- and cross-attention, as in LoFTR 

\subsubsection{Training Loss} The loss function is defined as $\mathcal{L} = \mathcal{L}_f + \mathcal{L}_c$, where $\mathcal{L}_f$ and $\mathcal{L}_c$ are fine- and coarse-level losses, respectively. We directly adopted the fine-level loss $\mathcal{L}_f$ of LoFTR \cite{sun2021loftr}. It considers $l_2$ loss of fine-level matches with the total variance on a cropped patch.

For coarse-level loss $\mathcal{L}_c$, we define a new loss function considering the topic model. Given a set of ground truth matches $\mathcal{M}_c$ at a coarse level, we label each ground truth pair as one. 
The loss for the positive samples has the following form.

\begin{multline}
    \mathcal{L}_c^{pos} = -\sum_{m_{ij} \in \mathcal{M}_c} \Bigl( E_{p(z_{ij})}  \log P(m_{ij} \mid z_{ij}, F_c^A, F_c^B) + \\ + \log \sum_{k=1}^K \theta_{i,k}^A \theta_{j,k}^B \Bigl)
\end{multline}
where the first term represents the ELBO loss estimated by Eqs. \ref{eq:conditional_match_dis} and \ref{eq:topic_sampling}, and the second term is used to enforce the pair on the same topic, which is derived from Eq. \ref{eq:topic_dis_all_k}.

We also needed to add a negative loss to prevent the assignment of all features to a single topic. For each ground truth match $m_{ij}$, we sampled $N$ unmatched pairs $\{m_{in}\}_{n=1}^N$ and then defined the negative loss using Eq. \ref{eq:topic_dis_nan}:

\begin{equation}
    \mathcal{L}_c^{neg} = - \sum_{m_{ij}} \left( \frac{1}{N} \sum_{n=1}^N \log \Bigl( 1 - \sum_{k=1}^K \theta_{i,k}^A \theta_{n,k}^B \Bigl) \right)
\end{equation} 
The final coarse-level loss involves these positive and negative terms, $\mathcal{L}_c = \mathcal{L}_c^{pos} + \mathcal{L}_c^{neg}$

\begin{table}[t]
\resizebox{8.5cm}{!}{%
  \centering
  \begin{tabular}{m{20em}|m{1.5em}m{1.5em}m{1.5em}|p{1.3em}}
    \Xhline{1.1pt}
    \multirow{2}{*}{\textbf{Method}} & \multicolumn{3}{p{5.4em}|}{\textbf{Homo. Est. AUC (\%)}} & \multirow{2}{*}{\textbf{\#M}} \\
    \cline{2-4}
    & 3px & 5px & 10px & \\
    \Xhline{1.1pt}
    \textbf{D2Net} \cite{dusmanu2019d2} + \textbf{NN} & 23.2 & 35.9 & 53.6 & 0.2K \\
    \textbf{R2D2} \cite{revaud2019r2d2} + \textbf{NN} & 50.6 & 63.9 & 76.8 & 0.5K \\
    \textbf{DISK} \cite{tyszkiewicz2020disk} + \textbf{NN} & 52.3 & 64.9 & 78.9 & 1.1K \\
    \textbf{SP} \shortcite{detone2018superpoint} + \textbf{SuperGlue} \shortcite{sarlin2020superglue} & 53.9 & 68.4 & 81.7 & 0.6K \\
    \textbf{Sparse-NCNet} \citep{rocco2020efficient} & 48.9 & 54.2 & 67.1 & 1.0K \\
    \textbf{DRC-Net} \cite{li2020dual} & 50.6 & 56.2 & 68.3 & 1.0K \\
    \textbf{Patch2Pix} \cite{zhou2021patch2pix} & 59.3 & 70.6 & 81.2 & 0.7K \\
    \textbf{LoFTR} \cite{sun2021loftr} & 65.9 & 75.6 & 84.6 & 1.0K \\
    \textbf{TopicFM} (Ours) & \textbf{67.3} & \textbf{77.0} & \textbf{85.7} & 1.0K \\
    \Xhline{1.1pt}
  \end{tabular}}
  \caption{Evaluation of \textbf{homography estimation} on HPatches \cite{balntas2017hpatches}. We compute AUC metrics following \citet{sun2021loftr}. \textbf{\#M} denotes the number of estimated matches}
  \label{tab:homo_hpatches}
\end{table}

\begin{table}[t]
\resizebox{8.5cm}{!}{%
  \centering
  \begin{tabular}{m{14em}|ccc}
    \Xhline{1.1pt}
    \multirow{2}{9em}{\textbf{Method}} & \multicolumn{3}{m{14em}}{\textbf{Relative Pose Estimation (MegaDepth / Scannet)}} \\
    \cline{2-4}
    & $5^o$ & $10^o$ & $20^o$ \\
    \Xhline{1.1pt}
    \textbf{SP} \shortcite{detone2018superpoint} +  \textbf{SuperGlue} \shortcite{sarlin2020superglue} & 42.2/16.16 & 61.2/\underline{33.81} & 76.0/\textbf{51.84} \\ 
    \textbf{DRC-Net}$^\star$ \cite{li2020dual} & 27.0/7.69 & 43.0/17.93 & 58.3/30.49 \\ 
    \textbf{Patch2Pix}$^\star$ \shortcite{zhou2021patch2pix} & 41.4/9.59 & 56.3/20.23 & 68.3/32.63 \\
    \textbf{LoFTR}$^\star$ \cite{sun2021loftr} & 52.8/\underline{16.88} & 69.2/33.62 & 81.2/50.62 \\ 
    \textbf{MatchFormer}$^\star$ \cite{wang2022matchformer} & \underline{52.9}/- & \underline{69.4}/- & \textbf{82.0}/- \\
    \textbf{TopicFM}$^\star$ (ours) & \textbf{54.1}/\textbf{17.34} & \textbf{70.1}/\textbf{34.54} & \underline{81.6}/\underline{50.91} \\
    \Xhline{1.1pt}
  \end{tabular}}
  \caption{Evaluation of \textbf{relative pose estimation} on MegaDepth and ScanNet. We use models trained only on MegaDepth for the coarse-to-fine methods denoted by $^\star$}
  \label{tab:rel_pose_megadepth}
\end{table}

\begin{figure}[ht]
\centering
\includegraphics[scale=0.35]{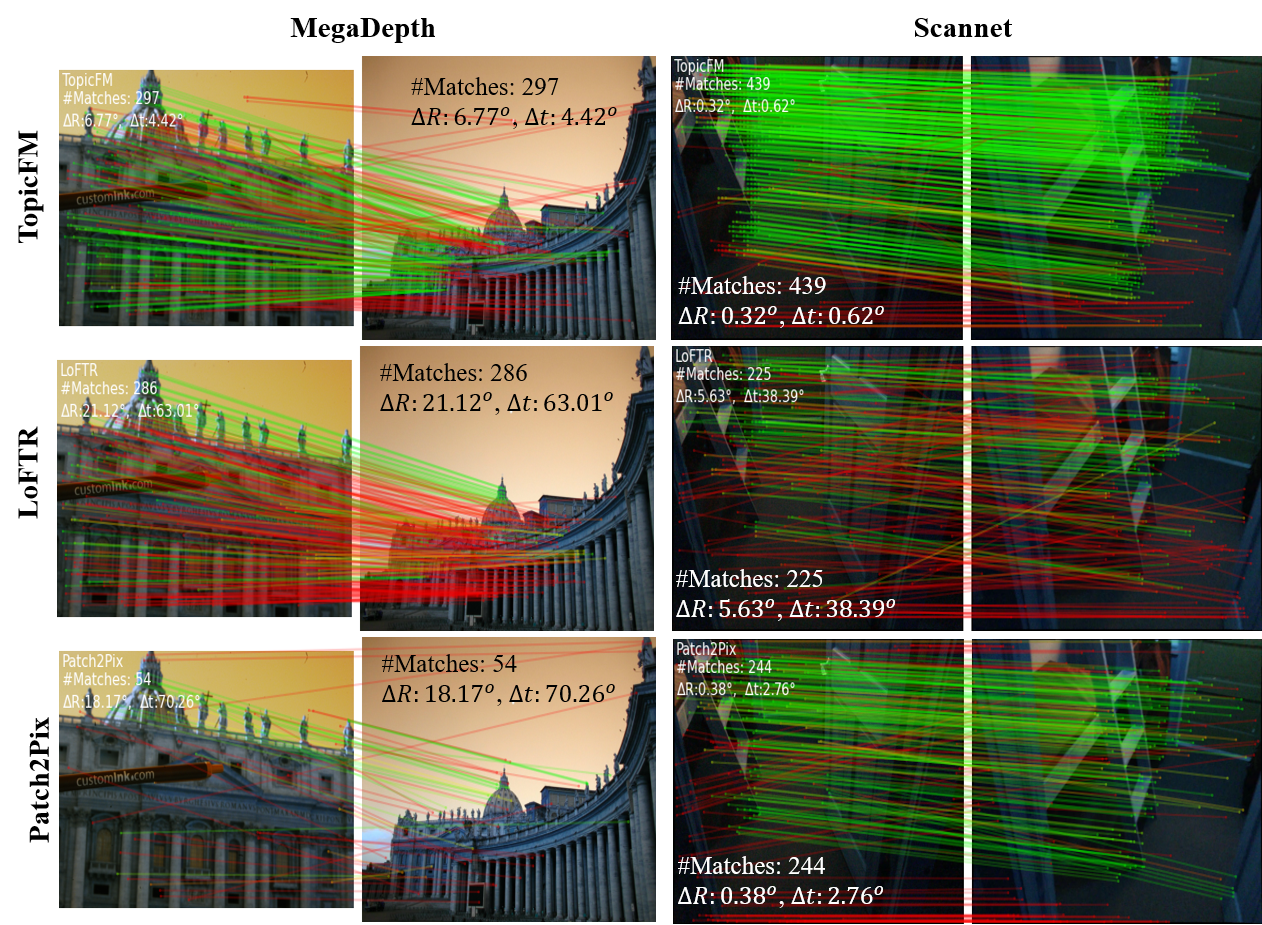}
\caption{Qualitative comparison between our method and other coarse-to-fine methods Patch2Pix and LoFTR. Our method can produce a high number of accurate correspondences in challenging conditions such as large relative viewpoints (MegaDepth) or untextured scenes (Scannet).}
\label{fig:rel_pose_qualitative}
\end{figure}

\begin{table}[t]
\resizebox{8.5cm}{!}{%
  \centering
  \begin{tabular}{p{13em}|cc|p{1.5em}}
    \Xhline{1.1pt}
    \multirow{2}{*}{\textbf{Method}} & \textbf{Day} & \textbf{Night} & \multirow{2}{1.5em}{\textbf{over- all}} \\
    \cline{2-3}
    & \multicolumn{2}{c|}{(0.25m,$10^o$)/(0.5m,$10^o$)/(1.0m,$10^o$)} & \\
    \Xhline{1.1pt}
    \textbf{ISRF} \cite{melekhov2020image} & 87.1/94.7/98.3 & 74.3/86.9/97.4 & 89.8 \\
    \textbf{KAPTURE} + \textbf{R2D2} + \textbf{APGeM} \cite{humenberger2020robust} & \underline{90.0}/\textbf{96.2}/\textbf{99.5} & 72.3/86.4/97.9 & 90.4 \\
    \textbf{SP} \shortcite{detone2018superpoint} + \textbf{SuperGlue} \shortcite{sarlin2020superglue} & 89.8/\underline{96.1}/\underline{99.4} & 77.0/\underline{90.6}/\textbf{100.0} & \underline{92.1} \\
    \textbf{Patch2Pix} \shortcite{zhou2021patch2pix} & 86.4/93.0/97.5 & 72.3/88.5/97.9 & 89.2 \\
    \textbf{LoFTR} \shortcite{sun2021loftr} & 88.7/95.6/99.0 & \textbf{78.5}/\underline{90.6}/99.0 & \underline{91.9} \\
    \textbf{TopicFM} (Ours) & \textbf{90.2}/95.9/98.9 & \underline{77.5}/\textbf{91.1}/\underline{99.5} & \textbf{92.2} \\
    \Xhline{1.1pt}
  \end{tabular}}
  \caption{Evaluation of \textbf{visual localization} on Aachen Day-Night v1.1 \cite{zhang2021reference}. We report the results using HLoc pipeline \cite{sarlin2019coarse}}
  \label{tab:vis_loc_aachen}
\end{table}
\begin{table}[t]
\resizebox{8.5cm}{!}{%
  \centering
  \begin{tabular}{p{14em}|cc|p{1.5em}}
    \Xhline{1.1pt}
    \multirow{2}{*}{\textbf{Method}} & \textbf{DUC1} & \textbf{DUC2} & \multirow{2}{1.5em}{\textbf{over- all}} \\
    \cline{2-3}
    & \multicolumn{2}{c|}{(0.25m,$10^o$)/(0.5m,$10^o$)/(1.0m,$10^o$)} & \\
    \Xhline{1.1pt}
    \textbf{ISRF} \cite{melekhov2020image} & 39.4/58.1/70.2 & 41.2/61.1/69.5 & 56.6 \\ 
    \textbf{KAPTURE} \cite{humenberger2020robust} + \textbf{R2D2} \cite{revaud2019r2d2} & 41.4/60.1/73.7 & 47.3/67.2/73.3 & 60.5 \\
    \textbf{SP} \shortcite{detone2018superpoint} + \textbf{SuperGlue} \shortcite{sarlin2020superglue} & \underline{49.0}/68.7/80.8 & 53.4/\textbf{77.1}/82.4 & 68.6 \\
    \textbf{Patch2Pix} \shortcite{zhou2021patch2pix} & 44.4/66.7/78.3 & 49.6/64.9/72.5 & 62.7 \\
    \textbf{LoFTR} \shortcite{sun2021loftr} & 47.5/72.2/84.8 & \underline{54.2}/\underline{74.8}/\textbf{85.5} & \underline{69.8} \\
    \textbf{CoTR} \shortcite{jiang2021cotr} & 41.9/61.1/73.2 & 42.7/67.9/75.6 & 60.4 \\
    \textbf{MatchFormer} \shortcite{wang2022matchformer} & 46.5/\underline{73.2}/\underline{85.9} & \textbf{55.7}/71.8/81.7 & 69.1 \\
    \textbf{TopicFM} (Ours)	& \textbf{52.0}/\textbf{74.7}/\textbf{87.4} & 53.4/\underline{74.8}/\underline{83.2} & \textbf{70.9} \\

    \Xhline{1.1pt}
  \end{tabular}}
  \caption{\textbf{Visual localization} on InLoc dataset \cite{taira2018inloc} using HLoc pipleline. We achieve best performance in overall.}
  \label{tab:vis_loc_inloc}
\end{table}

\section{Experiments}
\subsection{Settings and Datasets}

\textbf{Training} We trained the proposed network model on the MegaDepth dataset \cite{li2018megadepth}, in which the highest dimension of the image was resized to 800. Compared with state-of-the-art transformer-based models \cite{sarlin2020superglue,sun2021loftr} (e.g., LoFTR \shortcite{sun2021loftr} requires approximately 19GB of GPU), our model is much more efficient. Therefore, we used only four GPUs with 11GB of memory to train the model with a batch size of 4. We implemented our network model in PyTorch, with an initial learning rate of 0.01. For the network hyperparameters, we set the number of topics $K$ to 100, threshold of coarse match selection $\tau$ to 0.2, and number of covisible topics for feature augmentation $K_{co}$ to 6. 

We evaluated the image-matching performance on three application tasks: i) homography estimation, ii) relative pose estimation, and iii) visual localization. All of these experiments used the pre-trained model of MegaDepth without fine-tuning. However, some hyperparameters, including $\tau$ and $K_{co}$ can be modified during testing. 

\begin{figure*}[ht]
\centering
\includegraphics[scale=0.52]{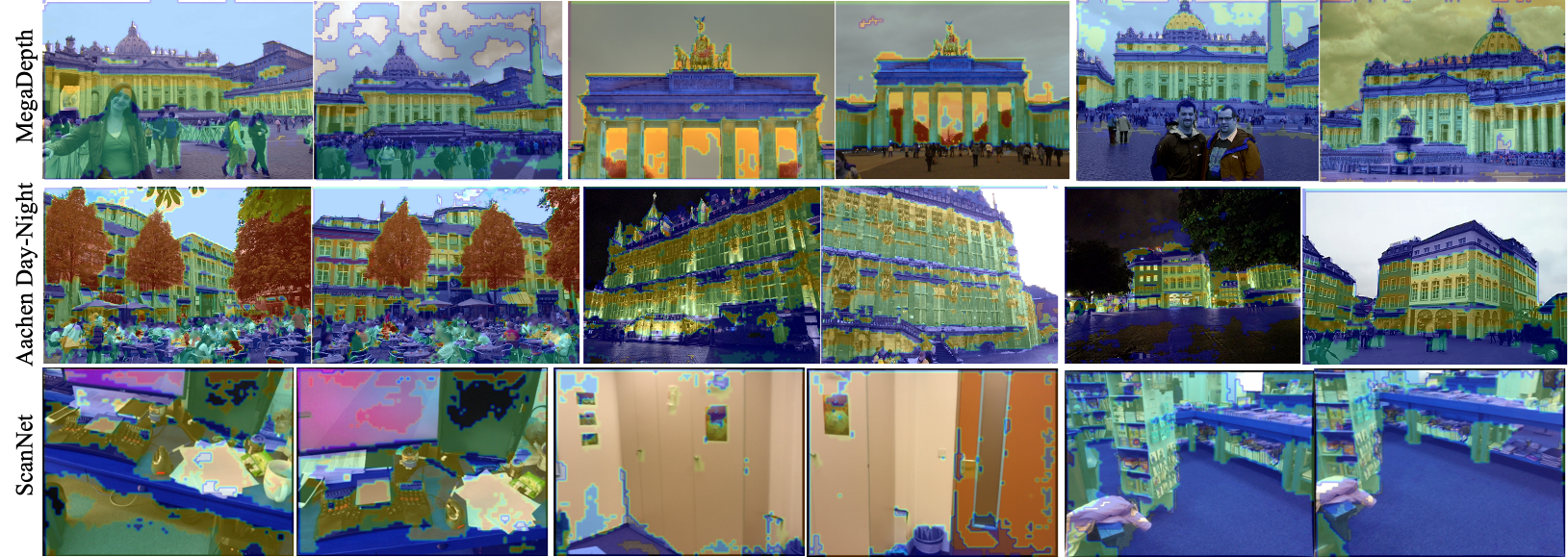}
\caption{Topic visualization across images and datasets. Our method can model a specific kind of structure by a topic that then supports the matching process effectively, as described in the method section.}
\label{fig:interpretability}
\end{figure*}

\subsection{Benchmark Performance}
\label{sec:benchmark}
\subsubsection{Homography estimation} The homography matrix between two images can be estimated by matching correspondences using the algorithm \cite{hartley2003multiple}. We used the HPatches dataset \cite{balntas2017hpatches} to estimate the homography matrices. For each image pair, we first warped the four corners of the first image to the second image based on the estimated and ground-truth homographies. We then computed the corner error between the two warped versions \cite{detone2018superpoint}. The error was measured using the AUC metric with thresholds of 3, 5, and 10 pixels \cite{sarlin2020superglue}. To report the results, we followed the same setup as in LoFTR. Table \ref{tab:homo_hpatches} shows the homography estimation performance of our method and the state-of-the-art methods. Our method generally outperformed other methods, demonstrating its effectiveness

\subsubsection{Relative pose estimation} 
To evaluate the image-matching performance, we measured the accuracy of the transformation matrix between the two images. We tested outdoor (MegaDepth \cite{li2018megadepth}) and indoor (Scannet \cite{dai2017scannet}) datasets. Each test set includes 1500 image pairs of images. We set the image resolution to $640\times480$ for Scannet and resized the highest dimension of the image to 1200 for MegaDepth. Similar to \cite{sarlin2020superglue,sun2021loftr}, we measured the area under the cumulative curve (AUC) of the pose estimation error at thresholds of $\{5^o,10^o,20^o\}$. 

Table \ref{tab:rel_pose_megadepth} shows the AUC results for both MegaDepth and ScanNet datasets. To make a fair comparison on ScanNet, we used models trained only on MegaDepth for all the coarse-to-fine methods. As shown in Table 2, our method performed better than the other coarse-to-fine baselines 
for all evaluation metrics. Compared with SuperPoint (SP) \cite{detone2018superpoint} + SuperGlue \cite{sarlin2020superglue}, our method had a worse performance only at $20^o$ of AUC on the Scannet. The main reason for this is that SuperGlue is trained directly on the Scannet. However, TopicFM was still better than SP+SuperGlue. We provide a detailed comparison with additional baselines for ScanNet in the Supplementary Material.

\subsubsection{Visual Localization} 
Unlike relative pose estimation, visual localization aims to estimate a camera pose for each image in a global coordinate system; however, it involves several steps. First, the pipeline builds a 3D structure of the scene from a set of database images. Next, given an input query image, it registers this image into the database and finds a set of 2D-3D matches that are then used to output the pose of the query image. Finding correspondences plays an important role in these steps. Therefore, we plugged the matching method into a visual localization pipeline to evaluate the matching performance. Following Patch2Pix, we use a full localization pipeline with HLoc \cite{sarlin2019coarse}. The benchmark datasets were the Aachen Day-Night v1.1 containing outdoor images and the InLoc dataset with indoor scenes. 

Tables \ref{tab:vis_loc_aachen} and \ref{tab:vis_loc_inloc} present the results for the Aachen v1.1 \cite{zhang2021reference} and InLoc \cite{taira2018inloc} datasets, respectively. Our method achieved competitive performance on both benchmarks compared with state-of-the-art baselines. As shown in Table \ref{tab:vis_loc_aachen}, TopicFM had a similar overall performance to SP+SuperGlue. SP and SuperGlue are trained by leveraging different types of datasets with various shapes and scenes, such as MS-COCO 2014 \cite{lin2014microsoft} (SP), synthetic shapes (SP), and MegaDepth (SuperGlue). Compared with the second-best LoFTR method, our overall result was slightly better. The main reason for achieving a satisfactory performance of LoFTR is that it was fine-tuned by augmenting the color images of MegaDepth to fit the nighttime images. In contrast to all the aforementioned setups, our method uses only a unified model trained on MegaDepth. This demonstrated the robustness of the proposed architecture. Similarly, for the InLoc evaluation shown in Table \ref{tab:vis_loc_inloc}, our method is better for all baselines on the DUC1 set with a large margin, although it is worse on the DUC2 set. However, we still achieved the best performance on average.

\subsection{Interpretability Visualization}
\label{sec:formatting}
We visualized the inferred topics to demonstrate the interpretability of the proposed model. As shown in Fig. \ref{fig:interpretability}, our method can partition the contents of an image into different types of spatial structures, in which the same semantic instances are assigned to the same topic. For instance, the topic “human” is marked in green color in the first image pair of MegaDepth and Aachen; the "tree" is marked in orange, and the "ground" is in blue. Different parts of a building, such as roofs, windows, and pillars, are separated into different topics. This phenomenon was repeated across images of MegaDepth and Aachen Day-Night, demonstrating the effectiveness of our topic modeling and inference modules. Notably, as illustrated in the third image pair of the first two rows in Fig. \ref{fig:interpretability}, our method focuses on the covisible structures in the same topic (marked with color) and ignores the non-overlapping information (marked without color). Although TopicFM was trained on the outdoor dataset MegaDepth, it could still generalize well on the indoor dataset ScanNet, as shown in the last row of Fig. \ref{fig:interpretability}. 


\section{Conclusion}
We introduced a novel architecture using latent semantic modeling for image matching. Our method can learn a powerful representation without high computational power by leveraging adequate context information in latent topics. As a result, the proposed method is robust, interpretable, and efficient compared with state-of-the-art methods.

\section{Acknowledgments}
This work was supported by the National Research Foundation of Korea (NRF) funded by the Ministry of Education under Grant 2016R1D1A1B01013573; and the Industrial Strategic Technology Development Program (No. 20007058, Development of safe and comfortable human augmentation hybrid robot suit) funded by the Ministry of Trade, Industry, \& Energy (MOTIE, Korea).


\bibliography{aaai23}

\clearpage
\appendix
This supplementary material provides additional information not covered in the main manuscript. We first provide the details of the proposed network architecture and experimental setups. We then provide several ablation studies, including the influence of the number of (i) topics and (ii) covisible topics and (iii) runtime comparison. Finally, we provide some additional qualitative results of matches on various challenging cases.

\section{Network Architecture Details}

\begin{table*}[t]
\resizebox{17.5cm}{!}{%
  \centering
  \begin{tabular}{|m{5em}|c|c|c|m{6em}|}
    \hline
    & Stage & TopicFM & LoFTR & Output size \\
    \hline
    \multirow{17}{5em}{\textbf{Feature Extraction}} & \multirow{2}{3em}{$F_1$} & $\text{Conv}[K:7,S:2,C:128]+\text{BN}+\text{ReLU}$ & $\text{Conv}[K:7,S:2,C:128]+\text{BN}+\text{ReLU}$ & \multirow{2}{15em}{$\frac{H}{2} \times \frac{W}{2} \times 128$} \\
     & & $[\text{Conv}[K:3,S:1,C:128]+\text{BN}+\text{GELU}]_{\times 2}$ & $[\text{ResBlock}[K:3,S:1,C:128]+\text{BN}+\text{ReLU}]_{\times 2}$ & \\
     \cline{2-5}
      & \multirow{2}{3em}{$F_2$} & $\text{Conv}[K:3,S:2,C:192]+\text{BN}+\text{GELU}$ & $\text{ResBlock}[K:3,S:2,C:196]+\text{BN}+\text{ReLU}$ & \multirow{2}{6em}{$\frac{H}{4} \times \frac{W}{4} \times 192$ $(196)$} \\
     & & $\text{Conv}[K:3,S:1,C:192]+\text{BN}+\text{GELU}$ & $\text{ResBlock}[K:3,S:1,C:196]+\text{BN}+\text{ReLU}$ & \\
     \cline{2-5}
     & \multirow{2}{3em}{$F_3$} & $\text{Conv}[K:3,S:2,C:256]+\text{BN}+\text{GELU}$ & $\text{ResBlock}[K:3,S:2,C:256]+\text{BN}+\text{ReLU}$ & \multirow{2}{6em}{$\frac{H}{8} \times \frac{W}{8} \times 256$} \\
     & & $\text{Conv}[K:3,S:1,C:256]+\text{BN}+\text{GELU}$ & $\text{ResBlock}[K:3,S:1,C:256]+\text{BN}+\text{ReLU}$ & \\
     \cline{2-5}
     & \multirow{2}{3em}{$F_4$} & $\text{Conv}[K:3,S:2,C:384]+\text{BN}+\text{GELU}$ & \multirow{2}{*}{$-$} & \multirow{2}{6em}{$\frac{H}{16} \times \frac{W}{16} \times 384$} \\
     & & $\text{Conv}[K:3,S:1,C:384]+\text{BN}+\text{GELU}$ & & \\
     \cline{2-5}
     & \multirow{3}{3em}{$F_c=F_3^{out}$} & $\text{Conv}[K:3,S:1,C:384](F_3) \oplus \text{Up}(F_4)$ & \multirow{3}{*}{$\text{Conv}[K:3,S:1,C:256]$} & \multirow{3}{6em}{$\frac{H}{8} \times \frac{W}{8} \times 256$} \\
     & & $\text{Conv}[K:3,S:1,C:256]+\text{BN}+\text{GELU}$ & & \\
     & & $\text{Conv}[K:3,S:1,C:256]$ & & \\
     \cline{2-5}
     & \multirow{3}{3em}{$F_2^{out}$} & $\text{Conv}[K:3,S:1,C:256](F_2) \oplus \text{Up}(F_3^{out})$ & $\text{Conv}[K:3,S:1,C:256](F_2) \oplus \text{Up}(F_3^{out})$ & \multirow{3}{6em}{$\frac{H}{4} \times \frac{W}{4} \times 192$ $(196)$} \\
     & & $\text{Conv}[K:3,S:1,C:192]+\text{BN}+\text{GELU}$ & $\text{Conv}[K:3,S:1,C:256]+\text{BN}+\text{LReLU}$ & \\
     & & $\text{Conv}[K:3,S:1,C:192]$ & $\text{Conv}[K:3,S:1,C:196]$ & \\
     \cline{2-5}
     & \multirow{3}{3em}{$F_f = F_2^{out}$} & $\text{Conv}[K:3,S:1,C:192](F_1) \oplus \text{Up}(F_2^{out})$ & $\text{Conv}[K:3,S:1,C:196](F_1) \oplus \text{Up}(F_2^{out})$ & \multirow{3}{6em}{$\frac{H}{2} \times \frac{W}{2} \times 128$} \\
     & & $\text{Conv}[K:3,S:1,C:128]+\text{BN}+\text{GELU}$ & $\text{Conv}[K:3,S:1,C:196]+\text{BN}+\text{LReLU}$ & \\
     & & $\text{Conv}[K:3,S:1,C:128]$ & $\text{Conv}[K:3,S:1,C:128]$ & \\
     \hline
     \multirow{2}{5em}{\textbf{Coarse Matching}} & \multirow{2}{5em}{TopicFM module ($T$-topics)} & $[\mathcal{CA}[C:256,\text{head}:8](T,F_c)]_{\times 5}$ & \multirow{3}{*}{$\left[ \begin{array}{c} \mathcal{SA}[C:256,\text{head}:8](F_c,F_c) \\ \mathcal{CA}[C:256,\text{head}:8](F_c^A,F_c^B) \end{array} \right]_{\times 4}$} & \multirow{2}{6em}{$\frac{H}{8} \times \frac{W}{8} \times 256$} \\
     \cline{3-3}
     & & $\left[ \begin{array}{c} \mathcal{SA}[C:256,\text{head}:8](F_c^k,F_c^k) \\ \mathcal{CA}[C:256,\text{head}:8](F_c^{A,k},F_c^{B,k}) \end{array} \right]_{k \in K_{co}}$ & & \\
     \hline
     \multirow{2}{5em}{\textbf{Fine Matching}} & & \multirow{2}{*}{$\mathcal{CA}[C:128,\text{head}:4]$} & $\mathcal{SA}[C:128,\text{head}:8]$ & \multirow{2}{6em}{$\frac{H}{2} \times \frac{W}{2} \times 128$} \\
     & & & $\mathcal{CA}[C:128,\text{head}:8]$ & \\
     \hline
    
  \end{tabular}}
  \caption{The detailed architectures of the proposed method (TopicFM) and the most related baseline (LoFTR). Our design improves the computational efficiency for all coarse-to-fine steps: i) feature extraction, ii) coarse matching, and iii) fine matching. Note that a ResBlock includes two convolutional blocks along with a residual connection. K, S, C, SA, and CA are denoted for the kernel size, stride, size of output channel, self-attention, and cross-attention, respectively.}
  \label{tab:network_arch_details}
\end{table*}

Table \ref{tab:network_arch_details} provides the details of the proposed network architecture. As mentioned in \textbf{Implementation Details} of the main paper, we design an adequate number of layers for each step to guarantee computational efficiency. We used fewer convolutional layers in the feature extraction than the baseline method, LoFTR \cite{sun2021loftr}. Moreover, the coarse-matching module of TopicFM is also more efficient than the multiple self/cross-attention blocks in LoFTR. In particular, the topic-aware attention of TopicFM is theoretically faster than the attention of LoFTR. When linear attention \cite{shen2021efficient} is applied, the computational complexity of TopicFM is $O(K_{co} \frac{N}{K})$, where $N$ is the number of image features. In contrast, the complexity of LoFTR is $O(N)$. Otherwise, if the standard dot-product attention \cite{dosovitskiy2020image} is employed, the complexity of TopicFM is $O(K_{co} \frac{N^2}{K^2})$ while the standard attention requires $O(N^2)$. Finally, the fine-matching module of TopicFM only applied cross-attention instead of self/cross-attention as in LoFTR. Therefore, our architecture is more efficient than LoFTR.

\subsubsection{Initial Feature Extraction} Given two images, $I^A$ and $I^B$, our method applies UNet-like CNN layers \cite{lin2017feature} with four down-up sampling steps to extract multi-scale dense feature maps. Similar to LoFTR, we utilize low-resolution feature maps $F_c^A$ and $F_c^B$ at $\frac{1}{8}$ scale for coarse matching and high-resolution feature maps $F_f^A$ and $F_f^B$ at $\frac{1}{2}$ scale for fine matching. We adopt a standard convolution block for feature extraction to enhance computational efficiency instead of the ResNet block as in LoFTR.

\subsubsection{Coarse Matching} In this step, our method takes the feature maps $F_c^A$ and $F_c^B$ as inputs and computes matching probabilities to determine their correspondences. To find the coarse matching, TopicFM first infers the topics for each feature point on $F_c^A$ and $F_c^B$ via Transformers, in which the topics are queries. A topic represents a hidden semantic structure like a latent instance or structural shape. The set of topics is globally shared across images in a dataset. Each image can be modeled as a multinomial distribution over topics, which is the basic concept of topic modeling in data mining \cite{blei2003latent,yan2013biterm}.

Next, we produce more informative features by fusing the global context information of topics with the local photometric context of $F_c^A$ and $F_c^B$. We then compute the matching probability for the informative features. The details of TopicFM are described in the method section of the main manuscript. Finally, we select pairs of feature points with high similarity as the coarse-matching correspondences by thresholding the matching probabilities. 

\subsubsection{Fine Matching} After determining the coarse matches, we refine them using the fine-level feature maps, $F_f^A$ and $F_f^B$. The match refinement method of LoFTR is directly adopted for fine matching. We upscale the coordinates of the coarse matches and crop a $N_p \times N_p$ patch from $F_f^A$ and $F_f^B$ centered at each coordinate. We then apply a cross-attention layer to enrich feature information inside each patch. For each pair of patches, we fix the center coordinate of the first patch and compute pair-wise similarity to the features of the second patch. Finally, we extract the refined matches from the pair of coordinates with the highest similarity.

\begin{figure}[h]
\centering
    \includegraphics[scale=0.21]{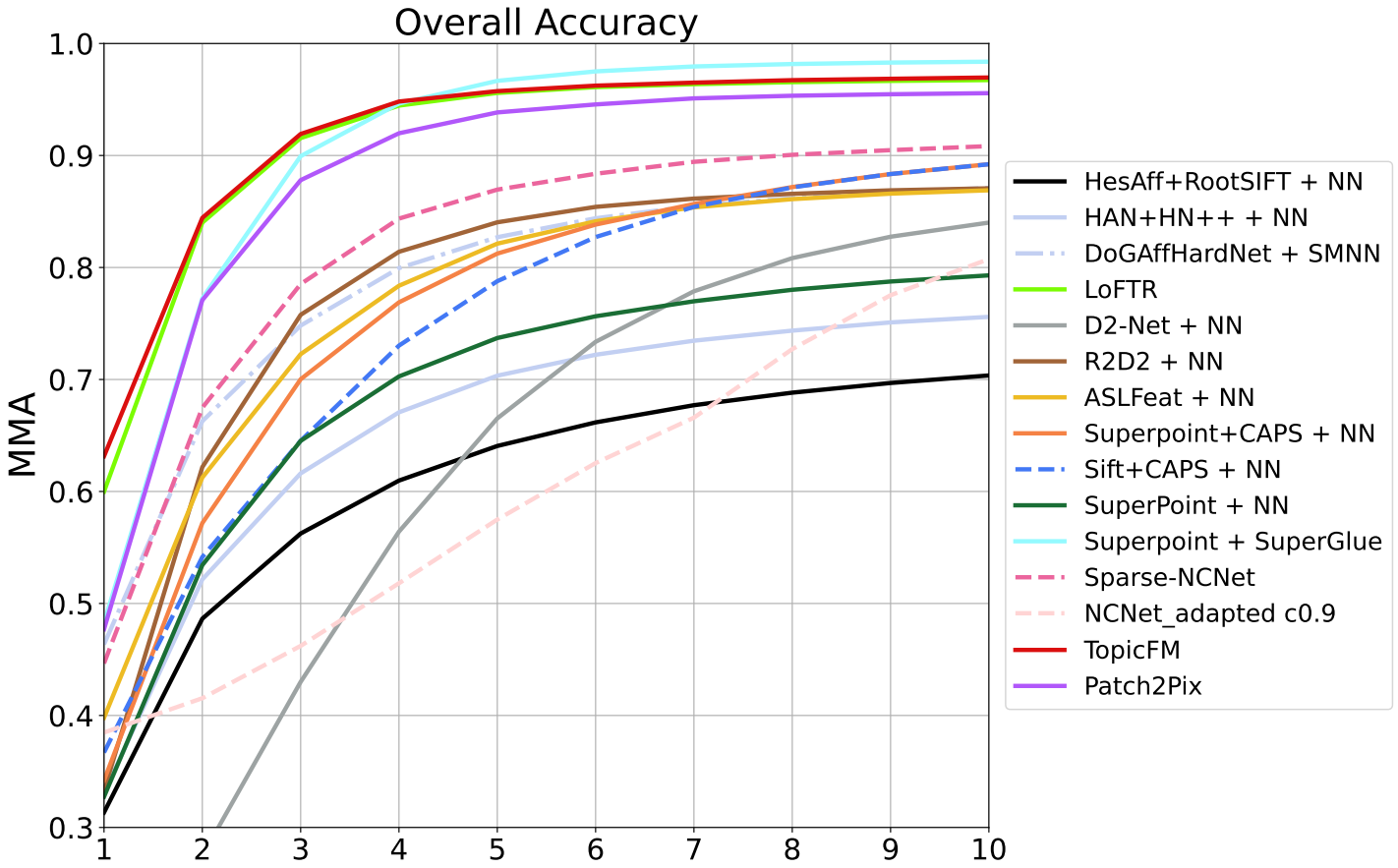}
    \caption{Mean Matching Accuracy (MMA) computed on HPatches. We report the results when changing thresholds from 1 to 10 pixels. TopicFM outperforms the others from [1,5] pixels.}
    \label{fig:mma_hpatches}
\end{figure}

\section{Experiment Details}
\subsection{HPatches Evaluation Details}
We provided the evaluation for homography estimation on the HPatches dataset in the main manuscript. In this evaluation, we rescaled the images so that the shorter dimension of each image is equal to 480. We set the number of covisible topics $K_{co}$ to 7 and the coarse-matching threshold $\tau$ to 0.2. Similar to LoFTR, we only selected the average of 1000 matches and then used the \textit{findHomography} function in OpenCV to estimate the homography matrices. We set the thresholds of RANSAC and confidence to 3 and 0.99999, respectively. 

This supplementary material also provides an additional evaluation of image matching on HPatches. We followed the same settings of Patch2Pix \cite{zhou2021patch2pix}, in which the longer dimension of each input image was rescaled to 1024. The model hyper-parameters $K_{co}$ and $\tau$ were set to 7 and 0.9, respectively. The results of all baselines were directly obtained by the image matching toolbox \cite{zhou2021patch2pix}. Running our method on HPatches only requires 8 GB of memory on an NVIDIA Geforce GPU. Fig. \ref{fig:mma_hpatches} shows the mean matching accuracy (MMA) for our method (TopicFM) and other state-of-the-art methods. As shown in Fig. \ref{fig:mma_hpatches}, our method generally achieves high accuracy in overall. In particular, our method showed the best performance in the range of 1 to 4 pixels.

\subsection{Detailed Evaluation on ScanNet}
\begin{table}[t]
\resizebox{8.5cm}{!}{%
  \centering
  \begin{tabular}{m{15em}|p{2.5em}p{2.5em}p{2.5em}}
    \Xhline{1.1pt}
    \multirow{2}{*}{\textbf{Method}} & \multicolumn{3}{c}{\textbf{Rel. Pose Est. (ScanNet)}} \\
    \cline{2-4}
    & $3^o$ & $5^o$ & $10^o$ \\
    \Xhline{1.1pt}
    ORB \cite{rublee2011orb} + GMS \cite{bian2017gms} & 5.21 & 13.65 & 25.36 \\
    D2-Net \cite{dusmanu2019d2} + NN & 5.25 & 14.53 & 27.96 \\
    ContextDesc \cite{luo2019contextdesc} + Ratio Test \cite{lowe2004distinctive} & 6.64 & 15.01 & 25.75 \\
    SP \shortcite{detone2018superpoint} + NN & 9.43 & 21.53 & 36.40 \\
    SP \shortcite{detone2018superpoint} + PointCN \cite{yi2018learning} & 11.40 & 25.47 & 41.41 \\
    SP \shortcite{detone2018superpoint} + OANet \cite{zhang2019learning} & 11.76 & 26.90 & 43.85 \\
    SP \shortcite{detone2018superpoint} + SuperGlue \shortcite{sarlin2020superglue} & 16.16 & \underline{33.81} & \textbf{51.84} \\
    DRC-Net$^\star$ \cite{li2020dual} & 7.69 & 17.93 & 30.49 \\
    Patch2Pix$^\star$ \shortcite{zhou2021patch2pix} & 9.59 & 20.23 & 32.63 \\
    LoFTR$^\star$ \cite{sun2021loftr} & \underline{16.88} & 33.62 & 50.62 \\
    TopicFM$^\star$ (ours) & \textbf{17.34} & \textbf{34.54} & \underline{50.91} \\
    \Xhline{1.1pt}
  \end{tabular}}
  \caption{The results of relative pose estimation on Scannet. To make a fair comparison, we used models trained only on MegaDepth for the detector-free methods (denoted as $^\star$).}
  \label{tab:rel_pose_scannet}
\end{table}

Table \ref{tab:rel_pose_scannet} presents more evaluation results of relative pose estimation on the Scannet dataset. We considered various state-of-the-art methods, including the detector-based and detector-free methods. As shown in Table \ref{tab:rel_pose_scannet}, our method outperforms all the detector-free methods trained on the MegaDepth dataset. For the detector-based methods, only the Superpoint (SP) + SuperGlue achieved a competitive performance because SuperGlue was trained directly on 230M image pairs of ScanNet.

\subsection{Visual Localization Details}
We evaluated the performance of visual localization task on two datasets: Aachen Day-Night v1.1 and Inloc. We followed Patch2Pix \cite{zhou2021patch2pix} to implement the full visualization pipeline of HLoc \cite{sarlin2019coarse}. The computed results of the camera pose were submitted to the visual localization benchmark \footnote{https://www.visuallocalization.net/benchmark} to obtain the AUC accuracy. Note that all results of baseline methods in Tables 4 and 5 of the main manuscript were also obtained from the benchmark webpage.

\subsubsection{Aachen Day-Night v1.1} This dataset includes 6697 database images and 1015 query images (824 day-time and 191 night-time images). To generate image pairs, HLoc selected the top-20 and top-50 nearest neighbors for each image in the database and query sets, respectively. We rescaled the longer image dimension into 1200 and set the hyperparameters $K_{co}$ and $\tau$ to 6 and 0.2, respectively. According to Patch2Pix, the correspondences were further refined by a quantization step. The keypoints whose pairwise distances are smaller than 2 pixels are merged into a single one. This step can improve the consistency of detected keypoints of the same image coming from the different image pairs. Finally, we ran the RANSAC with the threshold of 25 for pose estimation in HLoc. This pipeline requires about 30 hours to finish on an NVIDIA Tesla V100 with 32GB of GPU.

\subsubsection{InLoc} This indoor dataset includes 9972 database images and 329 query images. Because this dataset provides 3D database pointclouds, HLoc needs to generate the query image pairs by selecting the top-40 nearest retrieval neighbors for each query image. For our method, we set the longer dimension of an input image to 1024, the number of covisible topics $K_{co}$ to 8, and the coarse matching threshold $\tau$ to 0.1. We selected the maximum of 2048 highest confident matches as the final results. We then used a RANSAC threshold of 48 to estimate the camera pose. The total running time of the InLoc experiment is about two hours on an NVIDIA Tesla V100 with 32GB GPU.

\begin{table}[t]
\begin{minipage}[t]{1.0\linewidth}
\resizebox{8cm}{!}{%
  \centering
  \begin{tabular}{p{2em}|p{2.5em}p{2.5em}p{2.5em}||p{2em}|p{2.5em}p{2.5em}p{2.5em}}
    \Xhline{1.1pt}
    \multirow{2}{3em}{$K$} & \multicolumn{3}{c||}{\textbf{AUC on MegaDepth}} & \multirow{2}{3em}{$K_{co}$} & \multicolumn{3}{c}{\textbf{AUC on MegaDepth}} \\
    \cline{2-4} \cline{6-8}
    & $5^o$ & $10^o$ & $20^o$ & & $5^o$ & $10^o$ & $20^o$ \\
    \Xhline{1.1pt}
    10 & 48.6 & 65.4 & 78.3 & 2 & 49.3 & 66.1 & 78.6 \\
    20 & 49.1 & 65.3 & 77.8 & 6 & 52.7 & 69.0 & 81.1 \\
    50 & \textbf{49.1} & \textbf{66.2} & \textbf{78.7} & 8 & 53.3 & 69.6 & 81.2 \\
    80 & 49.1 & 65.7 & 78.1 & 10 & \textbf{54.1} & \textbf{70.1} & \textbf{81.6} \\
    100 & 49.0 & 65.6 & 77.9 & 12 & 53.7 & 69.8 & 81.4 \\
    \Xhline{1.1pt}
  \end{tabular}}
  \caption{Impact of the number of topics and covisibility-topics.}
  \vspace{0.5cm}
  \label{tab:ablation_topics}
\end{minipage}

\begin{minipage}[t]{1.0\linewidth}
\resizebox{8cm}{!}{%
  \centering
  \begin{tabular}{p{7em}|ccc}
    \Xhline{1.1pt}
    \multirow{2}{9em}{\textbf{Method}} & \multicolumn{3}{c}{\textbf{Runtime Benchmark (ms)}} \\
    \cline{2-4}
    & $640 \times 480$ & $896 \times 672$ & $1200 \times 896$ \\
    \Xhline{1.1pt}
    Patch2Pix & 228 & 659 & 1935 \\
    LoFTR & 100 & 232 & 500 \\
    TopicFM & \textbf{75} & \textbf{172} & \textbf{323} \\
    \Xhline{1.1pt}
  \end{tabular}}
  \caption{Efficiency analysis of coarse-to-fine methods.}
  \vspace{-0.3cm}
  \label{tab:ablation_runtime}
\end{minipage}
\end{table}

\section{Ablation Study}

\subsubsection{Number of Topics} This section analyzes the influence of the number of topics $K$ used in TopicFM. Table \ref{tab:ablation_topics} shows the results of relative pose estimation for TopicFM on MegaDepth when using $K \in \{10,20,50,80,100\}$. We used the same hyperparameters as the experiment of MegaDepth in the main manuscript and calculated AUC metrics. We trained each model of $K$ using 30\% of the training set to get results faster. We fixed the number of covisible topics $K_{co}$ to 6 and the coarse matching threshold $\tau$ to 0.2 in both training and test phase. As observed in Table \ref{tab:ablation_topics}, the model with $K=50$ achieved the best performance. However, the best performance was not significantly better than other models. Therefore, we used $K$ of 100 when training on the full MegaDepth dataset and evaluating the performances on the other datasets of our experiments. 

\subsubsection{Covisible Topics} TopicFM selects the most important topics by estimating the covisible probabilities of topics (refer to the \textbf{Topic-assisted Feature Matching} of the main manuscript). Therefore, the number of covisible topics $K_{co}$ also affects the overall matching performance. To analyze the influence of $K_{co}$, we evaluated the performance of TopicFM for different $K_{co}$ values. Table 8 shows the Relative Pose Estimation results on MegaDepth when using $K_{co} \in \{2,4,6,8,10,12\}$. As shown in Table \ref{tab:ablation_topics}, the performance is gradually increased when $K_{co}$ is increased, but it drops when $K_{co}$ is higher than 10. This indicates that considering many covisible topics could degrade the performance. Therefore, it is important to determine the appropriate number of covisible topics.

\begin{figure*}[h]
\centering
    \includegraphics[scale=0.38]{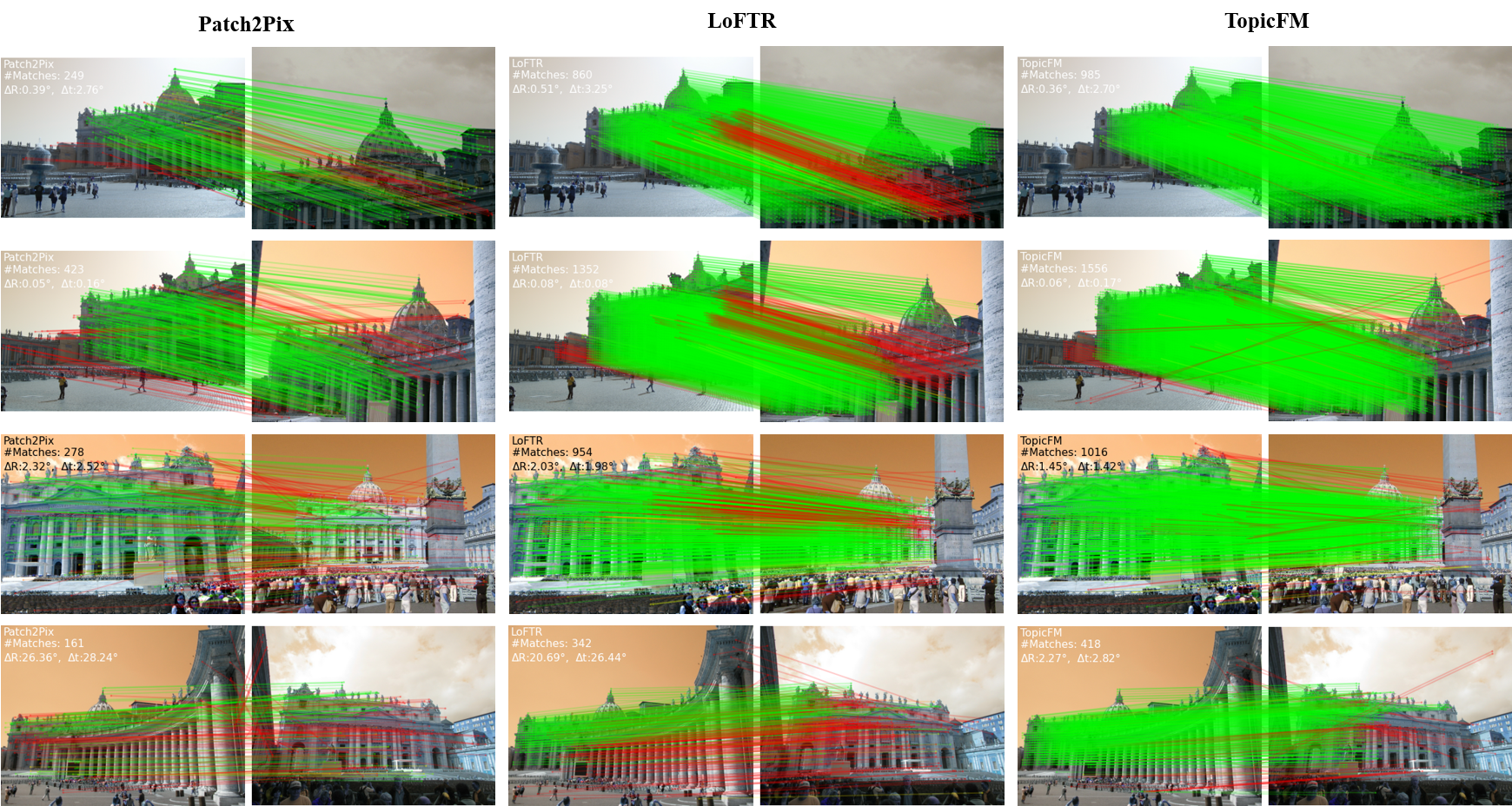}
    \caption{Qualitative evaluation of Patch2Pix \cite{zhou2021patch2pix}, LoFTR \cite{sun2021loftr} and our method (TopicFM) on MegaDepth dataset. Based on the known ground truth camera poses, we can visualize the correct matches in green color and the wrong matches in red colors. Our method produces a much higher number of matches compared to Patch2Pix and LoFTR. }
    \label{fig:megadepth_matches}
\end{figure*}

\begin{figure*}[p]
\centering
    \includegraphics[scale=0.6]{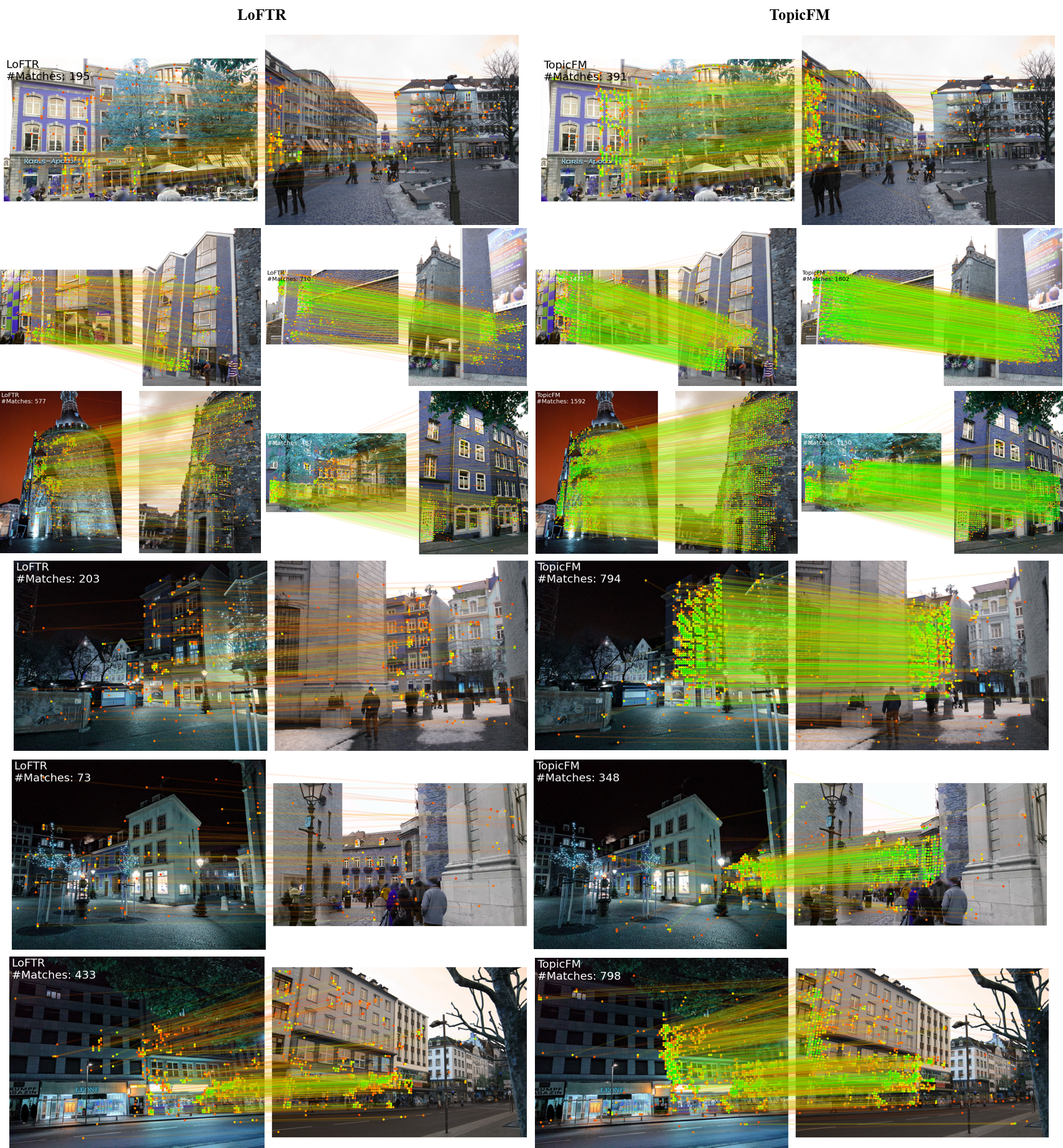}
    \caption{Qualitative results on the Aachen Day-Night v1.1 dataset. We selected both day-time and night-time image pairs for visualization. The color of matches is visualized based on the estimated matching confidence; the high-confidence matches are shown in green and low confidence in red. Our method produced a much higher number of high-confidence matches than LoFTR.}
    \label{fig:aachen_matches}
\end{figure*}

\begin{figure*}[p]
\centering
    \includegraphics[scale=0.58]{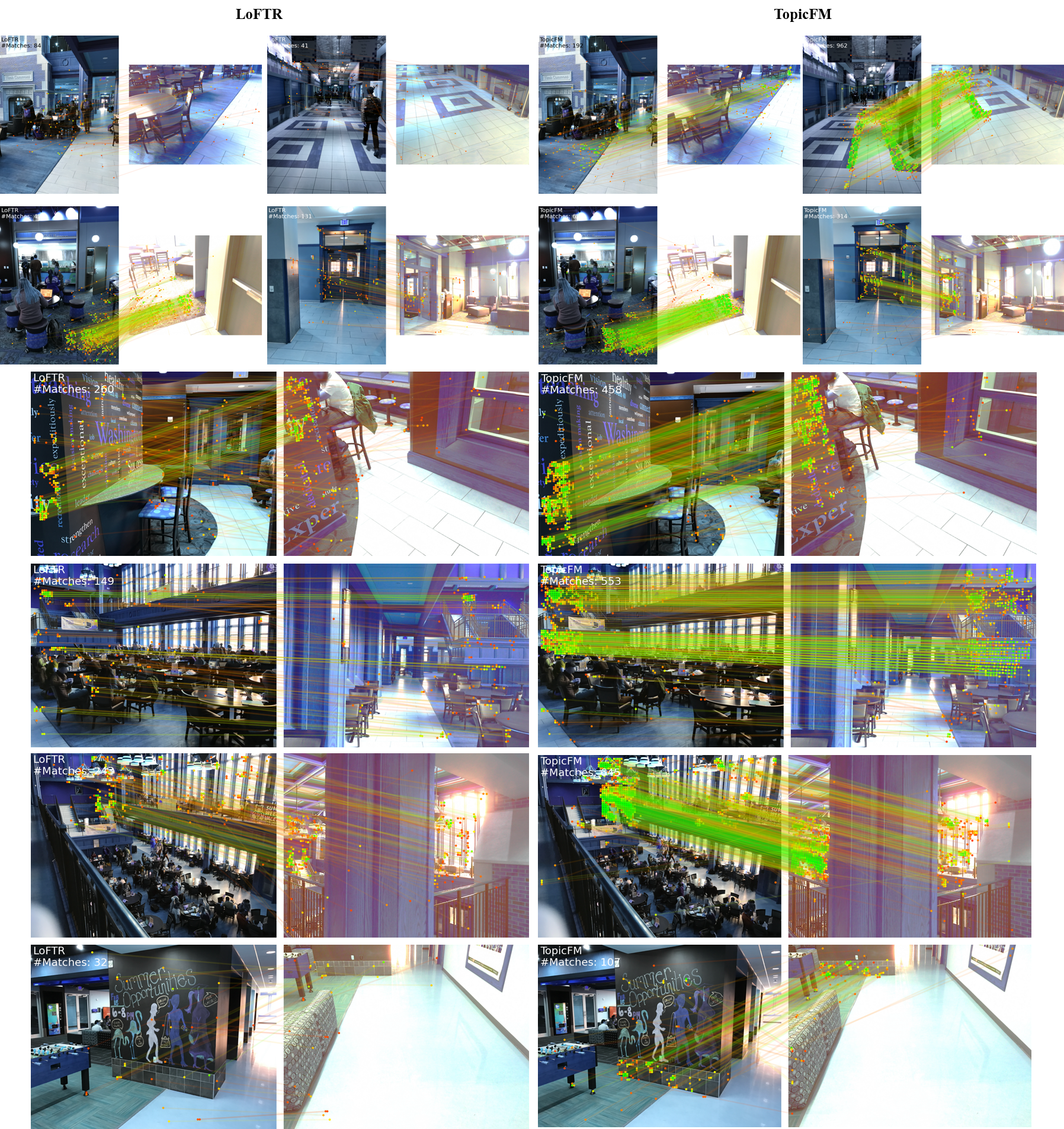}
    \caption{Qualitative results on the indoor dataset, InLoc. Our method outperforms LoFTR in some challenging conditions such as large relative viewpoints, repetitive structures, or illumination changes.}
    \label{fig:inloc_matches}
\end{figure*}

\subsubsection{Runtime comparison} We compared the efficiency of our method with the recent coarse-to-fine methods, Patch2Pix \cite{zhou2021patch2pix} and LoFTR \cite{sun2021loftr}. We measured the processing time of an image pair on a workstation with an Intel Xeon CPU (48 cores) with 252 GB of RAM and a Tesla V100 GPU with 32 GB of memory. Table \ref{tab:ablation_runtime} provides the average runtimes according to different image resolutions. TopicFM had the fastest runtime performance in all cases. At the highest resolution, $1200 \times 896$, TopicFM significantly reduces the runtime by about 35\% and 83\% compared to LoFTR and Patch2Pix, respectively. 

Our method was designed as an efficient, lightweight network for each coarse-to-fine step. The feature extraction network of our method is composed of a standard UNet instead of ResUnet, so it requires 95 GFLOPs at the image resolution of $640 \times 480$. FLOPs is the number of Floating point Operations used for measuring the complexity of a neural network, 1 GFLOPs = $10^6$ FLOPs. Meanwhile, the ResUnet of LoFTR requires 149 GFLOPs, which is higher than ours. Furthermore, our coarse matching module uses a single block of self/cross-attention and shares it across topics to extract features. Therefore, our coarse matching requires less computation of about 15 GFLOPs at the image resolution of 640×480 compared to 52 GFLOPs of LoFTR. 

\section{Additional Qualitative Results}
In this section, we provide additional qualitative results for various challenging cases. Fig. \ref{fig:megadepth_matches} shows the qualitative results of Patch2Pix, LoFTR, and TopicFM in several difficult matching conditions on the MegaDepth dataset. The red and green lines represent the wrong and correct matches, respectively. We also provide qualitative comparisons of TopicFM and LoFTR on the Aachen Day-Night v1.1 and InLoc datasets, shown in Figs. \ref{fig:aachen_matches} and \ref{fig:inloc_matches}. The datasets do not provide the ground truth camera poses of images. Therefore, we directly measured the matching confidence of detected matches and then color-coded the matches according to the confidence; the high-confidence matches are shown in green and low confidence in red. As shown in Figs. \ref{fig:megadepth_matches}, \ref{fig:aachen_matches}, and \ref{fig:inloc_matches}, our method produced a much higher number of matches than LoFTR and Patch2Pix. In particular, the high confidence matches were distributed over a large percentage in the results of our method.


\end{document}